\documentclass[11pt]{article}

\usepackage[preprint]{acl}

\usepackage{times}
\usepackage{latexsym}

\usepackage[T1]{fontenc}

\usepackage[utf8]{inputenc}

\usepackage{microtype}

\usepackage{inconsolata}

\usepackage{graphicx}

\usepackage{tabularx}
\usepackage{booktabs}
\usepackage{multirow}
\usepackage{svg}
\usepackage{tcolorbox}

\title{Evaluation of Multilingual Ability to Use\\Spatial Deictic Expressions in Vision-Language Models}

\author{
 \textbf{Kaito Watanabe\textsuperscript{1,2}},
 \textbf{Taisei Yamamoto\textsuperscript{1,2}},
 \textbf{Tomoki Doi\textsuperscript{1,2}},
 \textbf{Hitomi Yanaka\textsuperscript{1,2,3}}
\\
\\
 \textsuperscript{1} The University of Tokyo,
 \textsuperscript{2} Riken,
 \textsuperscript{3} Tohoku University
\\
 \small{
   \texttt{\{nglhdf,yamamo96,doi-tomoki701,hyanaka\}@is.s.u-tokyo.ac.jp}
 }
}

\begin{document}
\maketitle
\begin{abstract}
One of the expected abilities of vision-language models (VLMs) is spatial reasoning ability based on a given text and image.
To evaluate the spatial reasoning abilities of VLMs, we focus on the use of spatial deictic expressions, which are defined as spatial expressions whose referent is determined by their situational context, such as ``this'' and ``that''.
To handle spatial deictic expressions, VLMs must jointly reason over language and visual space, grounding context-dependent references in the image's spatial structure.
In addition, selecting appropriate spatial deictic expressions across languages requires VLMs to understand the language-specific spatial distinctions encoded by these expressions.
In this paper, we develop a benchmark\footnote{Our benchmark is available in \url{https://github.com/ynklab/multilingual-demonstratives-eval}} to evaluate the multilingual ability of VLMs to use spatial deictic expressions in four languages.
Our experiments using this benchmark reveal that the tested models use demonstratives in a manner different from that of humans, particularly in selecting the appropriate demonstratives based on the distance to the object.
\end{abstract}

\section{Introduction}
\label{chap:introduction}

In recent years, large language models (LLMs) and vision-language models (VLMs) have achieved remarkable progress due to their scalability.
Furthermore, a key feature of LLMs and VLMs is their ability to handle a wide range of tasks not only in English but also in multiple languages.
The development prompted vigorous attempts to evaluate their reasoning ability.
Previous studies have investigated the abilities of VLMs to capture spatial relations and spatial expressions, such as frames of reference\footnote{Frames of reference(FoR) are frameworks which are used to express the relative position of an object from the perspective of the other object.}~\citep{zhang2025visionlanguagemodelsrepresentspace,khemlani2025visionlanguagemodelsunreliable}.

Despite efforts, the ability of VLMs to utilize spatial deictic expressions, an important type of spatial expression, has not been explicitly studied. 
Deixis is the usage of expressions whose referent is dependent on the situation of utterance.
Spatial deictic expressions are expressions that depend on the space in which the utterance occurred, such as ``here'' or ``that''.
For example, suppose there is a pen in front of the speaker.
If the pen is far away, especially at an unreachable point, the speaker tends to say ``that pen'', while if it is near the speaker, the speaker may indicate it by ``this pen''.
As this example shows, deictic functions are fundamental to human language because they connect linguistic expressions with the physical environment.
Therefore, the evaluation of a VLM's proficiency in using spatial deictic expressions has linguistic significance for benchmarking its capacity for spatial reasoning.

We argue that the use of spatial deixis poses two major challenges for VLMs: cross-linguistic variation and inherent ambiguity.
First, to use spatial deictic expressions appropriately, multilingual VLMs need to understand their semantic differences across languages.
For example, while in English we use two demonstratives, proximal \footnote{Proximal demonstratives refer to the objects located closer to the speaker, while distal demonstratives refer to the objects located further.} ``this'' and distal ``that'', in Japanese we use three demonstratives, proximal \textit{kono}, distal \textit{ano}, and medial \textit{sono}.
As such, across languages, the number of kinds of spatial deictic expressions varies.
In addition, as in another example, although Spanish and Japanese both have three demonstratives: proximal, distal, and medial, the meaning of each word is not the same.
According to \citet{BA45166906}, Spanish medial \textit{ese} signifies a relatively intermediate place between the speaker and the addressee, while Japanese medial \textit{sono} indicates a place near the addressee.
Therefore, multilingual VLMs need to capture such subtle differences in the semantic nuances of spatial distances across languages to employ demonstratives in a manner similar to humans.
Second, \citet{coventry2023spatial}, a study investigating demonstrative usage among approximately 30 participants per language, reported that individuals do not necessarily use the same demonstrative in the same situation; there are individual differences in capturing the semantic nuances of spatial distance.
Thus, if multilingual VLMs succeed in learning spatial deixis expressions, they are expected to recognize differences across languages and reproduce the distribution of human performance in the use of demonstratives.

Based on these points, we developed a benchmark to measure the ability to use spatial deictic expressions in various languages, such as English and Japanese, and evaluated VLMs.
The task included in the benchmark is designed to investigate how the absolute distance of an object influences the choice of demonstratives of VLMs.
We also analyzed the performance of VLMs by comparing the results of experiments on human spatial deictic expressions reported in \citet{coventry2023spatial}.

The contributions of this paper are as follows:

\begin{enumerate}
    \item We constructed the first benchmark to evaluate the ability to employ demonstratives in VLMs.
    \item We revealed that open VLMs fail to use demonstratives in a human-like way, and the differences between humans vary across models.
    In particular, VLMs do not show the shift in the selection of demonstratives with distance, as observed in humans.
\end{enumerate}

\section{Background}
\label{chap:background}
\subsection{Analysis of Spatial Reasoning Ability in VLMs}
\label{sec:spatial-reasoning}

A large number of evaluations have been conducted on spatial reasoning ability in VLMs~\citep{liu2025spatialreasoningmultimodallarge}.
Those previous benchmarking works revealed that VLMs still lack sufficient human-level ability~\citep{khemlani2025visionlanguagemodelsunreliable,zhang-etal-2025-sphere}, especially for 3D world spatial reasoning tasks such as distance estimation~\citep{11094671,zhang-etal-2025-sphere} and positional relationships (e.g. \textit{front} or \textit{left})~\citep{liu2023visual,khemlani2025visionlanguagemodelsunreliable}.

Moreover, many benchmarks were developed only in English (the works cited in the preceding paragraph are all for English evaluation).
Only a few benchmarks contain tasks to evaluate the multilingual ability of spatial reasoning in VLMs~\citep{liu2024mmbenchmultimodalmodelallaround, haller2025pisabenchpisaindexmultilingual, zhang2025visionlanguagemodelsrepresentspace}.

\citet{liu2024mmbenchmultimodalmodelallaround} aims to evaluate the ability of VLMs robustly and holistically.
To realize this goal, the benchmark, MMBench, includes a task called ``spatial relationship'' that contains 2D spatial relationships, and ``physical relation'' that contains 3D spatial relationships.
The authors created a Chinese version of the benchmark, called MMBench-CN.
They demonstrated that most of the tested VLMs performed worse on MMBench-CN, but the gap in performance between English and Chinese for a model that achieves a high English score may be smaller.

\citet{haller2025pisabenchpisaindexmultilingual} constructed a benchmark called \textsc{PISA-Bench} by collecting questions from the PISA test, an international assessment of the academic performance of students.
It contains ``spatial and geometric reasoning'' tasks.
As the name implies, it contains a task that asks for the shape of an object viewed from behind.
As a result, they found a significant gap in performance between English and the others.

\citet{zhang2025visionlanguagemodelsrepresentspace} developed an evaluation protocol called COMFORT.
The authors focused on frames of reference (FoR) to evaluate the robustness, consistency, and flexibility of VLMs' spatial reasoning.
They constructed images of datasets using Blender~\citep{blender2018blender}.
The benchmark asks VLMs to answer whether one object has an indicated relationship to the other, for example, ``From the camera's viewpoint, is the ball behind the car?'' using an image of a ball and a car.
They concluded that VLMs lack the ability to use indicated FoR and have a bias to the English method of expression rather than a method specific to the language.

However, previous work does not focus on the extent to which VLMs handle spatial deictic expressions in spatial reasoning tasks.
In this paper, we analyze multilingual abilities to use spatial deictic expressions in VLMs.

\subsection{Spatial Deixis}
\label{sec:spatial-deixis}
\subsubsection{Deixis and Demonstratives}
\label{subsec:deixis}
\textbf{Deixis} is the usage of spatial expressions whose referent is dependent on the situation of utterance~\citep{BB19119305}.
Deictic expressions are actual expressions of deixis.
For example, given the sentence \textit{John is here.}, the word \textit{here} is a spatial deictic expression, so we cannot decide what the word \textit{here} refers to without considering the situational context of the utterance.
In this situation, the speaker doesn't need to explain the referent of \textit{Here}.
Instead, the listener has to determine the referent of \textit{here}.
In this way, the accurate recognition of the meaning of spatial deictic expressions requires precisely determining the referent of the word.

Deictic expressions are classified into several classes~\citep{BA45166906,fillmore1997lectures}, and spatial deictic expressions are one of them.
Spatial deictic expressions are deictic expressions whose referent is related to space.
In this paper, we focus on spatial deictic expressions, especially demonstratives.

\citet{coventry2023spatial} investigated how people with various native languages use different demonstratives.
In their experiments, most languages have two or three demonstratives.
The differences among demonstratives vary across languages, but, in general, they tend to depend on distance.
Especially, a demonstrative used to refer to nearby and faraway objects from the speaker is called \textit{proximal} and \textit{distal}, respectively.

Some languages also have another criterion to differentiate demonstratives.
For example, in Japanese language, \textit{ano} and \textit{sono} both refer to the object not near the speaker, but the difference depends on the distance from the addressee.
\citet{coventry2023spatial} revealed that this kind of dependency on distance from the addressee is also observed in some languages, such as Finnish or Korean.

In order to output spatial deictic expressions appropriately, multilingual VLMs have to recognize the differences in demonstratives among languages.

\subsubsection{Deixis and Language Models}
To the best of our knowledge, whether VLMs can handle spatial deictic expressions has not been sufficiently explored.
Existing attempts to evaluate the deixis understanding ability of language models are prone to focus on discourse deixis, a kind of deixis, because LLMs do not have access to external visual contextual information.
A benchmark called PUB~\citep{sravanthi-etal-2024-pub} evaluates the pragmatic reasoning abilities of LLMs, including understanding of deixis.
Their results show that almost all tested LLMs perform worse than humans, and the authors conclude that the poor scores are not due to world knowledge but to a lack of pragmatic reasoning.
While several studies of deixis and language models exist in the field of human-computer interaction~\citep{10910098,10670510}, there are few studies of evaluation in vision-language models.

\section{Method}
\label{chap:method}
To construct our benchmark, we refer to the linguistic experiments called ``memory game''~\citep{gudde2018spatial}.
A detailed description of the "memory game'' is provided as follows.

\subsection{Memory Game}
\label{sec:memory_game}
The memory game is a method to investigate the usage of demonstratives of humans without subjects noticing that linguistic investigation is conducted \citep{gudde2018spatial}\footnote{In the original paper \citep{gudde2018spatial}, another version of the memory game is also suggested, called ``memory version''.}.
This method aims to elicit demonstratives from subjects of the experiment in experimentally controllable and naturalistic situations.
To ensure the naturalistic setting, the experiment is presented as a memory experiment, and the purpose of investigating demonstrative usage is concealed from participants until the experiment is completed.

The procedure of the memory game in \citet{gudde2018spatial} is as follows.
First, subjects sit on one side of a long desk and are told that the experiments examine the influence of language on memory.
Then they are instructed to name the shape drawn on the disk, which is placed on the long desk, by pointing their finger at the disk.
When naming the object, the subjects must use three words: a demonstrative, a color, and the shape of the object, like ``that black cross''.
Experimenters record which demonstrative is used, and one trial of the experiment is done.
Experiments are conducted multiple times with changes in some variables, such as the disk's location.
The places of the disk are classified into three ``regions'' by distance, region 1 for 25-75 cm, region 2 for 100-150 cm, and region 3 for 175-225 cm.
The result of the experiment showed that these parameters have an influence on demonstrative use.

\citet{coventry2023spatial} conducted a memory game across various languages, and they analyzed how the usages of demonstratives are different among languages.
In their research, experiments were conducted with 29 languages (e.g., Japanese and English) and 874 subjects whose native language was one of them.
The memory game experiment is conducted by using a white disk with a shape drawn on it.
The variables are places, shapes, colors, and the addressee's location.
Places of the disk are classified into three ``regions'' by distance, like \citet{gudde2018spatial}, but they set regions in a slightly different way: region 1 for 25-50 cm, region 2 for 150-175 cm, and region 3 for 275-300 cm.
Shape is a shape drawn on the disk (e.g, a star, a cross, and a triangle).
Color is the color of the shape.
Through extensive experiments, they showed that native speakers use different demonstratives (distal, medial, and proximal) depending on the distance to the target.
Furthermore, they found that the usage patterns of these demonstratives across distances vary across languages; for example, Japanese speakers tend to use medial demonstratives (\textit{sono}) for objects at an intermediate distance, whereas Korean speakers use distal ones (\textit{jeo}). 

The memory game enables the evaluation of human demonstrative use across multiple languages, accounting for subtle ambiguities that arise depending on the distance between the speaker and the object.
In this study, we construct a benchmark for evaluating demonstrative use in VLMs based on the memory game paradigm.

\subsection{Task Setting}
\label{sec:task_setting}
Based on the memory game, we composed a VQA task to analyze the extent to which VLMs capture spatial deixis across languages.
We chose four languages for evaluation, Japanese, Korean, English, and Chinese, in which there were differences with respect to the usage of spatial deictic expressions.
There exist three demonstratives in Japanese (proximal \textit{kono}, distal \textit{ano} and medial \textit{sono}) and Korean (proximal \textit{i}, distal \textit{jeo} and medial \textit{geu}), and two demonstratives in English (proximal \textit{this}, and distal \textit{that}) and Chinese (proximal \textit{zh\`{e} ge}, and distal \textit{n\`{a} ge}).

We follow the task setting used in the memory game.
In each image (see \autoref{fig:sample}), one white disk with a colored shape is put on the black desk.
VLMs are instructed to describe the shape drawn on the disk, using a fixed format consisting of a demonstrative, a color, and the shape, such as ``this red circle.''

\subsection{Benchmark Construction}
\label{sec:benchmark_construction}

We constructed the datasets with Blender~\citep{blender2018blender} following the construction process of previous benchmarks~\citep{khemlani2025visionlanguagemodelsunreliable,zhang2025visionlanguagemodelsrepresentspace}, because we can easily tweak the location of objects and the brightness with it.
To be consistent with experimental settings regulated in \citet{gudde2018spatial}, we produced images with the aligned settings.
For example, we used a long desk without any patterns and did not place any objects around it, which can help to estimate the distance of the object.
We also tweaked the camera height or the room brightness so that VLMs could obtain sufficient information to answer.

\begin{figure}[t]
  \centering
  \begin{minipage}{0.3\textwidth}
    \centering
    \includegraphics[width=\textwidth]{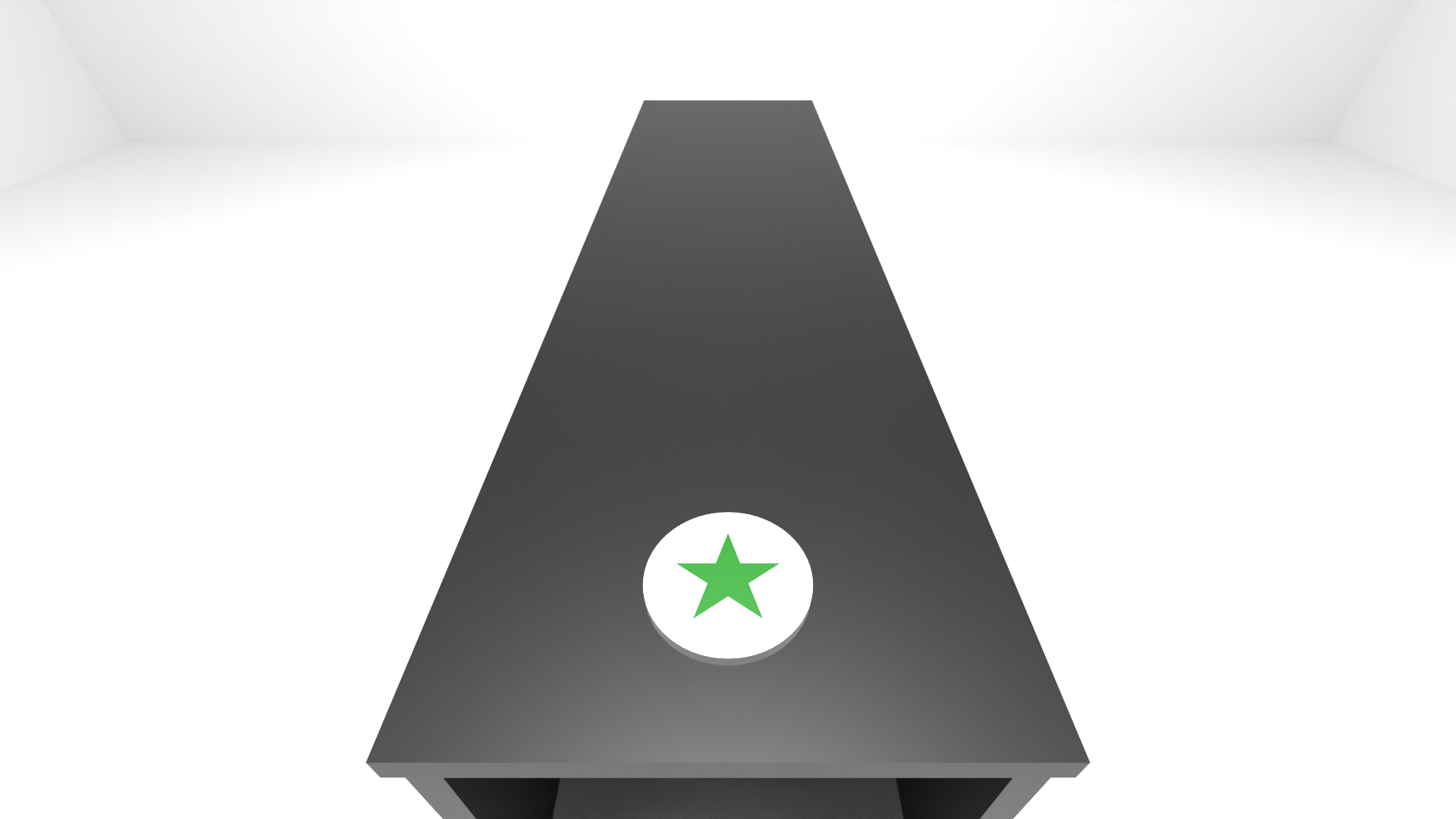}
  \end{minipage} \\
  \vspace{10pt}
  \begin{minipage}{0.3\textwidth}
    \centering
    \includegraphics[width=\textwidth]{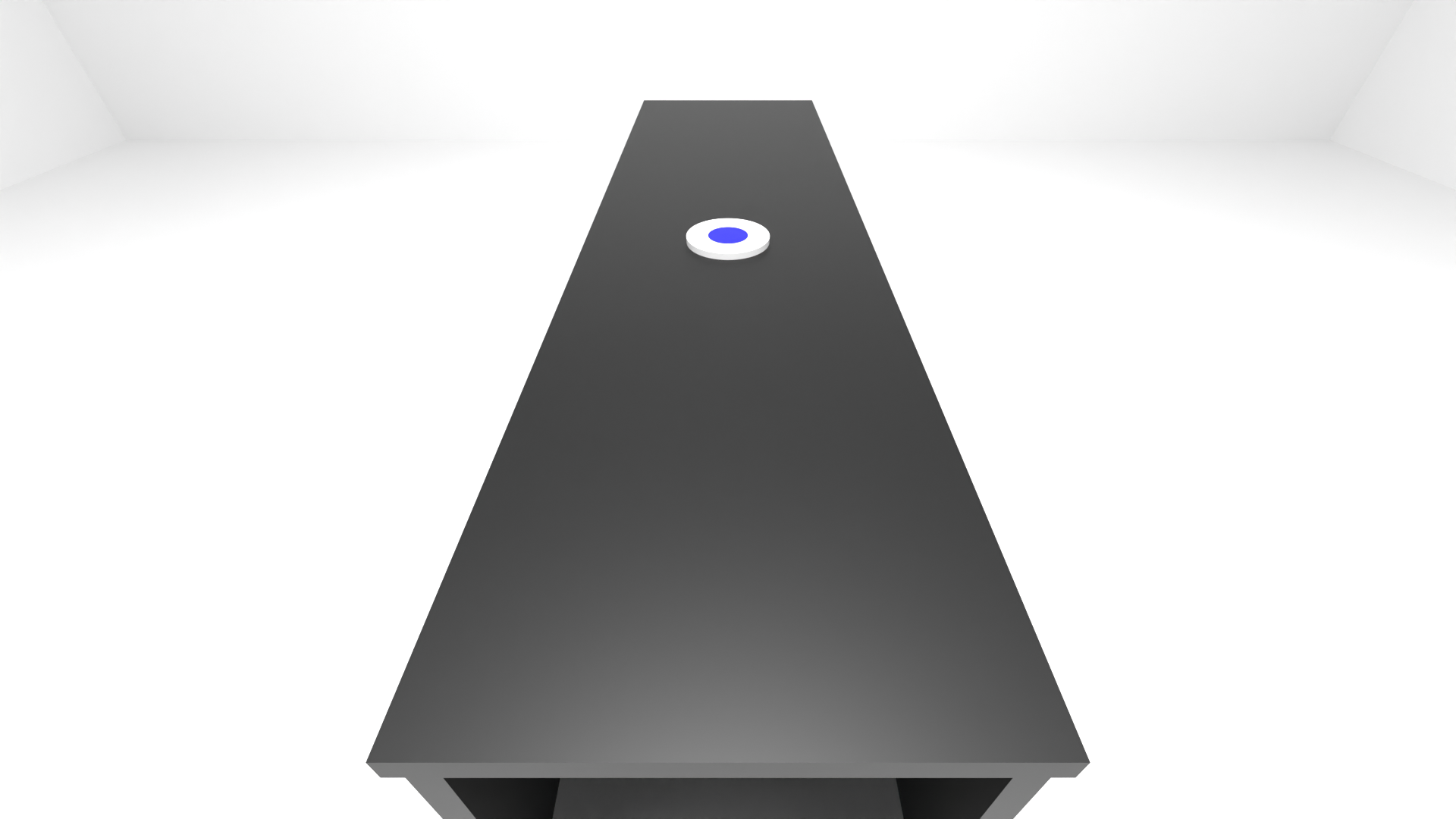}
  \end{minipage} \\
  \vspace{10pt}
  \begin{minipage}{0.3\textwidth}
    \centering
    \includegraphics[width=\textwidth]{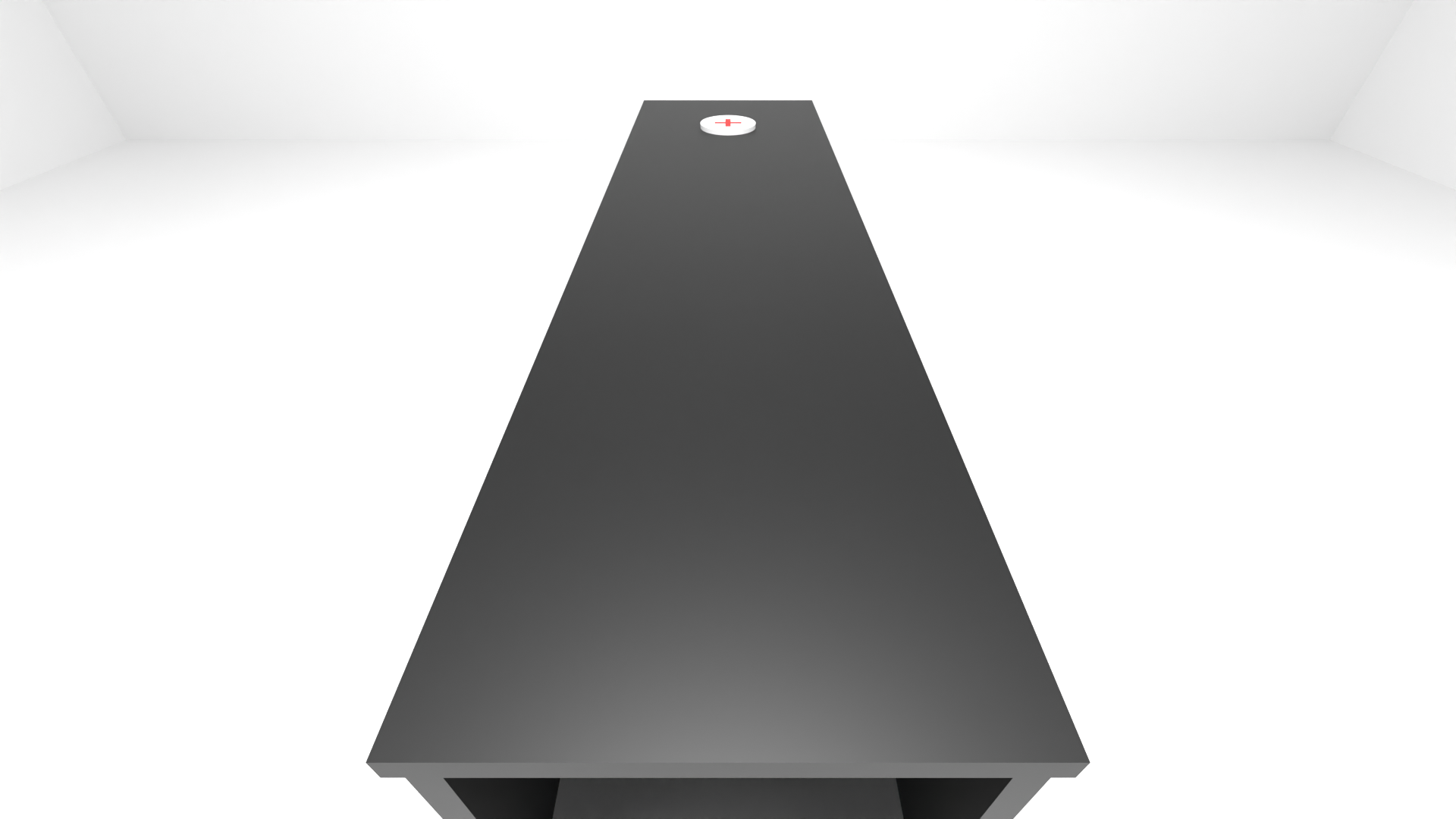}
  \end{minipage}
  
  \caption{Examples of images included in the benchmark. Distance from the object is 0.25m, 1.50m, and 2.75m, from left to right.}
  \label{fig:sample}
\end{figure}

Examples of the image included in the benchmark are displayed in \autoref{fig:sample}.
On the desk, we put a disk with a solid-colored shape.
The disk was placed at one of the distances 0.25 m, 1.50 m, and 2.75 m from the subject, as displayed in \autoref{fig:sample}, corresponding to region 1, 2, and 3 in \citet{coventry2023spatial} respectively.
We used 5 figures (circle, cross, square, star, and triangle) and 4 colors (black, blue, green, and red) to create 20 images per region, resulting in 60 images in total (5 shapes $\ times$4 colors for the shape $\times$ 3 positions for the disk).

\section{Experimental Setting}
\label{chap:experiments}
\subsection{Model}
\label{sec:models}
We chose Gemma 3 4B~\footnote{\url{https://huggingface.co/google/gemma-3-4b-it}} and Gemma 3 12B~\footnote{\url{https://huggingface.co/google/gemma-3-12b-it}} from Gemma 3 series~\citep{gemma_2025}, and Qwen3-VL 8B~\footnote{\url{https://huggingface.co/Qwen/Qwen3-VL-8B-Instruct}} and Qwen3-VL 32B~\footnote{\url{https://huggingface.co/Qwen/Qwen3-VL-32B-Instruct}} from Qwen3-VL series~\citep{Bai2025Qwen3VLTR} for evaluation.
All of them are open, instruction-tuned, and multilingual VLMs.
We avoided closed models such as GPT-5.2 because we cannot obtain raw logits for analysis from their APIs.
We accessed trained models of them through Hugging Face Hub\footnote{\url{https://huggingface.co/}}.
Output token decoding was performed using a greedy algorithm for all models.

\subsection{Prompt}
\label{sec:prompt}
Since the original instructions used in \citet{coventry2023spatial} experiments are not publicly available, we design prompts for VLMs following the protocol described in \citet{coventry2023spatial}.
We configured a prompt to have VLMs output the same format as the memory game for humans: a demonstrative, the color of the object, and a shape drawn on the disk placed on the long desk.
The prompt used in our experiments is shown below.

{
  \centering
  \begin{tcolorbox}
Prompt: Analyze the image and identify the shape on the disk. Describe it by filling in the following three-word template exactly:

[Demonstrative] [Color] [Shape]

Constraints:

Use "This" or "That" for the demonstrative.

Use a single word for the color.

Use a single word for the shape.

Output only the three words. Do not include a period or any introductory text.

  \end{tcolorbox}
}

This prompt was written in English at first, and then translated into each language by Google Gemini\footnote{\url{https://gemini.google.com/}}.
After the translation, we added lacking demonstratives, especially medial demonstratives of Japanese and Korean, because the literal translation of the prompt contained only proximal and distal demonstratives.

\subsection{Metrics}
\label{sec:metrics}
We primarily employ two metrics.
One is a probability distribution of the use of demonstratives for each VLM.
The other is the Jensen-Shannon distance between the probability distributions of demonstratives by models and humans.
This allows us to analyze how precisely models reproduce human usage of the spatial deictic expressions.

The probability distribution of the use of demonstratives is calculated from the logits obtained from outputs of LLMs.
We first calculate the logits of all target demonstratives, and then normalize them by applying the softmax function.
For example, we calculate the probability for ``this'' in English as $P(this) / (P(this) + P(that))$.

The fidelity of humans' selection of demonstratives is calculated referring to the probability distribution of humans reported on \citet{coventry2023spatial}.
We regarded the results of \citet{coventry2023spatial}, which were conducted for humans, as the probability distribution of demonstratives inherent in each language.
We measured the Jensen-Shannon distance between the distributions of humans and VLMs.
We adopted the Jensen-Shannon distance because it can measure the distance for a distribution that has a class whose probability is 0.

We calculated the Jensen-Shannon distance using the following formula:

\begin{equation}
JSD(P||Q) = \sqrt{\frac{D_{KL}(P||M)+D_{KL}(Q||M)}{2}}
\end{equation}

\noindent where $P$ and $Q$ is a probability distribution, $M$ is $\frac12 (P + Q)$ (pointwise mean), and $D_{KL} (P||M)$ denotes the Kullback-Leibler divergence from $M$ to $P$, defined as:

\begin{equation}
\displaystyle D_{KL} (P||M) = \sum_{d} P(d) \log \frac{P(d)}{M(d)}
\end{equation}

\noindent for each $d$ in demonstratives of the target language.

We compute the Jensen–Shannon distance between the distribution of demonstrative usage by the VLM and that by humans for each distance to the target, and compute a representative value for the model by averaging these distances across target distances.

Before the evaluation, we eliminated the inappropriate results that do not align with the format ``[Demonstrative] [Color] [Shape]'' and made some mistakes on colors and shapes, because VLMs fail to understand our instructions or to recognize the indicated object in those trials.
As colors and shapes can be expressed in many ways (e.g. ``cross'' and ``plus'' for a cross shape), we permitted those varieties of expressions during the validation.
The judgment to determine whether the output corresponds to the synonym of the gold answer was performed manually by the author.

\section{Results and Discussion}
\label{chap:results_and_discussion}
\subsection{Results and Analysis}
\label{sec:results_and_analysis}

\begin{figure*}[thbp]
  \centering
  \includesvg[height=293pt, keepaspectratio]{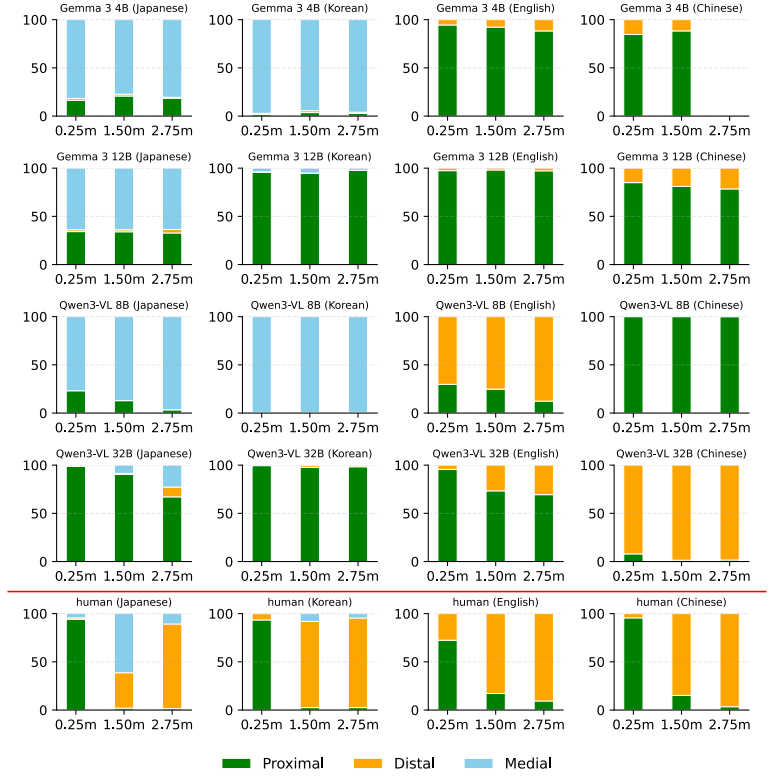}
  \caption{Probability distributions across distances for each experimental setting. The red horizontal line separates results for VLMs (above) from those for humans (below). Columns represent different languages, while rows above the red horizontal line correspond to specific VLMs. Human results in the bottom row were calculated based on the results in \citet{coventry2023spatial}.}
  \label{tab:result1}
\end{figure*}

\begin{table}[ht]
\centering
\resizebox{\columnwidth}{!}{
\begin{tabular}{l|cccc}
\hline
Models & Japanese & Korean & English & Chinese \\
\hline
Gemma 3 4B
 & 12 & 18 & 34 & 12 \\
\hline
Gemma 3 12B
 & 18 & 19 & 45 & 28 \\
\hline
Qwen3-VL 8B
 & 44 & 38 & 50 & 48 \\
\hline
Qwen3-VL 32B
 & 22 & 26 & 49 & 44 \\
\hline
\end{tabular}
}
\caption{The number of images in which the color and shape of the object are precisely recognized by the model (out of 60 images).}
\label{tab:table_recognition}
\end{table}

\begin{table*}[t]
\centering
\small
\begin{tabularx}{0.7\textwidth}{XX|c|c|c|c}
\hline
 Models & Distance(m) & Japanese & Korean & English & Chinese \\
\hline
\multirow{4}{*}{Gemma 3 4B}
 & 0.25 & 0.7227 & 0.9554 & 0.2615 & 0.1597 \\
 & 1.50 & 0.4591 & 0.8349 & 0.6830 & 0.6571 \\
 & 2.75 & 0.8252 & 0.8880 & 0.7174 & \textbackslash \\
 & Average & 0.6080 & 0.8705 & 0.5041 & \textbackslash \\
\hline
\multirow{4}{*}{Gemma 3 12B}
 & 0.25 & 0.5732 & 0.2067 & 0.3167 & 0.1557 \\
 & 1.50 & 0.5104 & 0.9173 & 0.7641 & 0.5839 \\
 & 2.75 & 0.7904 & 0.9367 & 0.8252 & 0.7018 \\
 & Average & 0.5815 & 0.7313 & 0.6326 & 0.5063 \\
\hline
\multirow{4}{*}{Qwen3-VL 8B}
 & 0.25 & 0.6749 & 0.9849 & 0.3677 & 0.1495 \\
 & 1.50 & 0.4821 & 0.8771 & 0.0799 & 0.8221 \\
 & 2.75 & 0.8510 & 0.9143 & 0.0413 & 0.9437 \\
 & Average & 0.6648 & 0.9200 & 0.1539 & 0.6499 \\
\hline
\multirow{4}{*}{Qwen3-VL 32B}
 & 0.25 & 0.1294 & 0.1638 & 0.2799 & 0.8175 \\
 & 1.50 & 0.8424 & 0.9130 & 0.4940 & 0.2277 \\
 & 2.75 & 0.7489 & 0.9247 & 0.5462 & 0.0510 \\
 & Average & 0.5055 & 0.7446 & 0.4444 & 0.3725 \\
\hline
\hline
\multirow{4}{*}{Uniform distribution}
 & 0.25 & 0.5740 & 0.5869 & 0.1962 & 0.4626 \\
 & 1.50 & 0.3907 & 0.5186 & 0.2998 & 0.3216 \\
 & 2.75 & 0.5149 & 0.5551 & 0.3912 & 0.4791 \\
 & Average & 0.4932 & 0.5535 & 0.2957 & 0.4211 \\
\hline
\end{tabularx}
\caption{Distance from human distribution of demonstrative use per language.}
\label{tab:result2}
\end{table*}

The number of images whose color and shape the model could precisely recognize is presented in \autoref{tab:table_recognition}, and the probability distributions of demonstratives for each VLM per language and distance settings are shown in \autoref{tab:result1}.
The probability distributions shown in \autoref{tab:result1} are calculated based only on the cases where images were correctly recognized by models, as presented in \autoref{tab:table_recognition}.
Each row of \autoref{tab:result1} corresponds to an experimental setting and presents the average probability distribution of demonstrative for that setting.
The bottom row is the result of the human investigations reported in \citet{coventry2023spatial}.
Green, orange, and light blue represent the probabilities of proximal, medial, and distal demonstratives, respectively.

Across models, a general trend is to avoid distal demonstratives in languages with three demonstratives, regardless of the model.
Although the proportion of distal demonstratives increased as the distance from the object increased for humans, the proportions of distal demonstratives in Korean and Japanese, which have three demonstratives, are consistently under 5\% except for the case of Qwen3-VL 32B for Japanese language.
Due to unbalanced distributions, models often fail to select the same demonstratives as humans do, especially in Japanese and Korean.

Regarding the trends for each model, Gemma 3 exhibits similar probability distributions across distances.
Although the proportion of proximal demonstratives decreases and that of distal demonstratives increases as the distance increases for humans, changes in the probability distribution of both Gemma 3 4B and Gemma 3 12B across different distances are smaller than the results of humans.
The Qwen3-VL series does not show such irrelevance to distance, but it fails to capture the human-like probability distribution.

\autoref{tab:table_recognition} presents that VLMs could not recognize the figures painted on the object and the content of the prompt in some experimental settings.
This phenomenon is particularly prominent in Gemma 3 4B, which exhibits biased distributions in demonstrative usage.
In contrast, Qwen3-VL achieves higher recognition accuracies than Gemma 3 models across all languages.
This may be caused by differences in the pre-training dataset.
While the pre-training dataset of Gemma 3 is not disclosed, that of Qwen3-VL includes tasks of visual question-answering with camera-captured images~\citep{Bai2025Qwen3VLTR}.
The fact that Qwen3-VL was pre-trained with similar tasks to ours may cause the more accurate recognition of the object in Qwen3-VL.
We can also observe that with Qwen3-VL 8B and 32B, English and Chinese scores of accurate recognition are higher than Japanese and Korean ones.
This can also be explained by the fact that it was pre-trained with data in which English and Chinese account for a large proportion~\citep{Bai2025Qwen3VLTR}.

It is also observable that while human preference for demonstratives varies according to distance, VLMs do not exhibit such sensitivity.
For instance, in Japanese, the most frequently selected demonstratives for humans shifted from proximal \textit{kono} at 0.25 m to medial \textit{sono} at 1.50 m, and finally to distal \textit{ano} at 2.75 m.
But Qwen3-VL 32B chose the proximal demonstrative at the highest proportion regardless of distance, and the other three models chose the medial at the highest proportion regardless of distance.
This lack of the shift of demonstratives by distance may be a feature of VLMs.

\subsection{Distance from Human Distribution}
\label{sec:distance-from-humans-distribution}

Jensen-Shannon distances between the outputs of models and humans' results of \citet{coventry2023spatial} are presented in \autoref{tab:result2}.
For comparison, the Jensen-Shannon distribution between the human distribution and the uniform distribution, in which every demonstrative is chosen with the same probability, is provided.
Diagonal lines in the table indicate that we cannot calculate the distance due to insufficient data resulting from a violation of the output format.

\autoref{tab:result2} shows Gemma 3 4B performs more differently from human distribution than uniform distribution in any language.
In contrast, Qwen3-VL 32B shows a similar distribution to humans in Japanese and Chinese, therefore, we can observe that Qwen3-VL 32B uses demonstratives in a relatively close way to humans.

We can also observe that the average distance in English is consistently smaller than that in Korean.
This can be explained in two ways.
First, Korean is a relatively low-resource language compared to English, and the shortage of training resources for Korean demonstratives may lead to low fidelity to humans' distribution.
Second, the probability distribution of the use of demonstratives in English is closer to a uniform distribution than that in Korean.
Thus, it is a relatively easier task to answer in a closer distribution in English than in Korean.

These results suggest that the lack of general object recognition ability in VLMs may cause biased demonstrative usage, even when color and shape are accurately recognized at evaluation time.
To further analyze the relationship between model object recognition ability and demonstrative usage, we calculated the correlation between image recognition ability and the distance of the probability distribution across 15 experimental settings, encompassing various combinations of languages and model architectures.
The analysis yielded a Pearson correlation coefficient of $r = -0.40$, indicating a modest negative trend between the two metrics.
This result suggests that models with higher recognition proficiency tended to exhibit demonstrative usage distributions more similar to those of humans.

\section{Conclusion}
\label{chap:conclusion}

In this paper, we developed a multilingual benchmark to assess the extent to which VLMs can use spatial deictic expressions in a manner similar to humans based on the memory game paradigm.
Using our benchmark, we investigated the probability distributions of demonstratives by object distance.

We discovered that all tested models fail to reproduce humans' probability distribution for demonstratives as a function of the absolute distance from the viewpoint to the target object.
We also observed that, unlike humans, some VLMs tend not to adjust their probability distributions based on distance to the referent.
These results suggest that current VLMs have significant room for improvement in their use of demonstratives in a human-like way.
We also clarified the difference in performance among languages.
In particular, all tested models show weaknesses in Korean language.
We also found that VLMs seldom use distal demonstratives in Japanese and Korean.

The results described above suggest that there are differences between VLMs and humans in handling spatial deictic expression, which is one of the fundamental expressions for spatial understanding.
We believe that our benchmark serves as a new testbed for evaluating the spatial reasoning abilities of VLMs across languages.

\section{Limitations}
\label{chap:limitation}

Though this benchmark introduces a novel methodology to evaluate the ability to use spatial deictic expressions, it is subject to several limitations.
First, the models cannot recognize a number of pictures, resulting in a constrained number of samples for evaluation.
Furthermore, the scope of our experimental setup is currently restricted.
Since our evaluation relies on a single prompt, it is difficult to rigorously determine whether the observed trends reflect the models' intrinsic capabilities or are artifacts of the prompt's phrasing.
The generalizability of these results is also constrained by the limited size of the dataset used and the limited number of languages and model architectures evaluated in this study.
Increasing the dataset size in future studies is required to mitigate potential biases and ensure a more robust assessment of the results.
In addition, since we needed to use a controlled prompt setting to ensure the VLM outputs satisfy the expected format (demonstrative, color, and shape), the variety of usage of demonstratives in natural situations is not sufficiently reflected.
Establishing more naturalistic settings remains as future work.
Finally, the evaluation process involved human judgment to verify model outputs.
It introduces an element of subjectivity and limits the benchmark's scalability.

\section*{Acknowledgments}

This work was supported by JST CREST Grant Number JPMJCR2565, Japan.
We would also like to thank Anirudh Reddy Kondapally for his valuable comments and suggestions.
The authors would also like to thank the anonymous reviewers for their constructive feedback, which helped to improve the quality of this paper.

\bibliography{custom,anthology-1,anthology-2}

\end{document}